\documentclass[runningheads]{llncs}
\usepackage{graphicx}
\usepackage{amsmath}
\usepackage{todonotes}
\usepackage{booktabs}
\usepackage{caption}

\usepackage[colorlinks=true, allcolors=blue]{hyperref}
\begin{document}
\title{Using Saliency and Cropping to Improve Video Memorability}

% 12 content pages, incl figures, tables, additional 2 pages containing only cited references, double blind
%
\titlerunning{Saliency and Cropping to Improve Video Memorability}
%\author{First Author\inst{1}\orcidID{0000-1111-2222-3333} \and
%Second Author\inst{2,3}\orcidID{1111-2222-3333-4444} \and
%Third Author\inst{3}\orcidID{2222--3333-4444-5555}}
%
\authorrunning{A.N. Author et al.}

\author{Vaibhav Mudgal\inst{1} \and Qingyang Wang\inst{1} \and Lorin Sweeney\inst{2}\orcidID{0000-1111-2222-3333} \and Alan F. Smeaton\inst{2}\orcidID{0000-0003-1028-8389}}
\authorrunning{Mudgal et al.}
\institute{School of Computing, Dublin City University, Ireland
\and
Insight Centre for Data Analytics, Dublin City University, Ireland\\
\email{alan.smeaton@dcu.ie}}

%\author{A.N. Author}
%\institute{Anon. University}
%\email{anonymous@mail.edu}

\maketitle              % typeset the header of the contribution
\begin{abstract} 
Video memorability is a measure of how likely a particular video is to be remembered by a viewer when that viewer has no emotional connection with the video content. It is an important characteristic as videos that are more memorable are more likely to be shared, viewed, and discussed. This paper presents  results of a series of experiments where we improved the memorability of a video by selectively cropping frames based on image saliency. We present results of a basic fixed cropping as well as the results from dynamic cropping where both the size of the crop and the position of the crop within the frame, move as the video is played and saliency is tracked. Our results indicate that especially for videos of low initial memorability, the memorability score can be improved. 

\end{abstract}
%-----------------------------------------------------------

\section{Introduction}
The saturation of contemporary society with digital content has rendered the ability to capture and sustain human attention increasingly elusive. In this replete landscape, the concept of ``video memorability” surfaces as a valuable construct. At its core, video memorability is commonly defined as encapsulating the propensity of a viewer to recognise a video upon subsequent viewing \cite{cohendet2019videomem,sweeney2022overview,sweeney2023diffusing}, a phenomenon that transcends the boundaries of emotional biases or personal connections. This assertion may appear counter-intuitive, given the prevailing inclination to associate memory with emotional resonance or subjective biases. However, the cognitive processes underlying memorability reveal it as an emotionally impersonal cognitive metric, intrinsically woven into the fabric of the content itself \cite{bainbridge2017memory}, and hence, impervious to the viewer's unique emotional landscape or individual predilections.
This conceptualisation of video memorability as an emotionally impersonal entity underscores the notion that certain videos innately possess characteristics that enhance their likelihood of being remembered, irrespective of the viewer's cognitive milieu. This intriguing facet of human cognition necessitates a more profound investigation, as it not only elucidates the complexities of our cognitive machinery, but also bears significant implications for a multitude of domains, including content creation, digital marketing, and education, among others.
An exploration of the literature reveals a paucity of research specifically dedicated to video memorability manipulation, despite a sizeable body of work on its corollary, image memorability. This disparity is, in part, attributable to the inherent complexities associated with videos, which, unlike static images, encompass dynamic spatial-temporal information. This additional layer of complexity engenders a host of challenges that have yet to be elegantly surmounted. Ultimately, a deeper appreciation of video memorability holds the promise of not only advancing our understanding of the human mind but also revolutionising the way we create, consume, and interact with digital content in an increasingly digital world.

This paper presents the findings from a series of experiments designed to augment memorability in short, 3-second videos, using a technique of selective frame cropping guided by visual saliency. Defined as the distinctiveness of certain elements or areas within an image or video frame that capture human visual attention, visual saliency \cite{cong2018review} serves as a cornerstone for our exploration into diverse cropping strategies. These range from fundamental fixed cropping to the more nuanced dynamic cropping, which not only adjusts the dimensions of the cropping frame but also its position, aligning with the shifts and movements of saliency throughout the video.

%In the next section we provide some background to the topic including an overview of the MediaEval task on predicting video memorability and on the dataset we use for our experiments. We then describe our experiments and the different approaches to cropping we have investigated. That is followed by our experimental results and conclusions.

\section{Background}

The characteristics
of images that make them more or less memorable than others were first explored from a computational viewpoint more
than a decade ago in \cite{isola2011understanding}. That work opened up a new domain for  researchers to
explore the field of image memorability and why some still images
are more memorable than others. The work in \cite{isola2011understanding} posited that the
memorability of an image is an intrinsic property and it aimed
at finding  visual attributes  that make an image
more, or less, memorable than others. It was found that images with people in
them tend to be more memorable than those without and that image memorability further
depends upon more precise attributes such as the peoples' ages,
hair colour and clothing.

Driven primarily by the MediaEval Media Memorability tasks \cite{sweeney2022overview,savran2021overview}, computational memorability has since evolved to encompass more complex visual stimuli, such as videos. In 2018, a video memorability annotation procedure was established, and the first ever large video memorability dataset---10,000 short soundless videos with both long-term and short-term memorability scores---was created \cite{cohendet2019videomem}. The integration of deep visual features with semantically rich attributes, such as captions, emotions, and actions, has been identified as a particularly efficacious strategy for predicting video memorability \cite{azcona2019ensemble,newman2020multimodal,sweeney2020leveraging,reboud2020predicting}. This confluence of modalities not only amplifies prediction precision but also furnishes a comprehensive perspective on the myriad factors that collectively shape video memorability. Furthermore, dimensionality reduction has been shown to enhance prediction outcomes, and certain semantic categories of objects or locales have been found to be intrinsically more memorable than others \cite{cohendet2019videomem}.

Building upon this foundation, recent research has further expanded the multifaceted approach by incorporating auditory features into the prediction model. A study by \cite{sweeney2021influence} illuminated the contextual role of audio in either augmenting or diminishing memorability prediction in a multimodal video context, thereby accentuating the necessity of a holistic approach that integrates diverse modalities, including visual, auditory, and semantic features, for more robust and accurate memorability predictions.

Moreover, the practical applications of video memorability prediction have been explored in real-world scenarios. A study by \cite{cummins2022analysing} conducted a comprehensive analysis of the memorability of video clips from the Crime Scene Investigation (CSI) TV series. Utilising a fine-tuned Vision Transformer architecture, the study predicted memorability scores based on multiple video aspects, dissecting the relationship between the characters in the TV series and the memorability of the scenes in which they appear. This analysis also probed the correlations between the significance of a scene and its visual memorability, revealing intriguing insights into the nexus between narrative importance and visual memorability.

Despite these advancements, the manipulation of video memorability remains a relatively uncharted territory. Existing literature predominantly focuses on predicting memorability scores rather than actively modifying them. This gap in the research underscores the need for novel approaches that not only predict but also enhance video memorability. The current study aims to address this gap by exploring the potential of saliency-based cropping as a technique for manipulating video memorability. By selectively cropping frames based on visual saliency, we hypothesise that it is possible to highlight the most memorable regions of a video, thereby enhancing its overall memorability.

\subsection{The MediaEval Benchmarking Task on Predicting Video Memorability}

In recent years, much of the work on computational prediction of the memorability of short form videos has been brought together as a task within the annual MediaEval benchmarking activity. Each year the organisers of the task share a collection of videos among participants who are asked to compute and submit runs which predict the memorability of each video in the collection. Once runs are submitted, the organisers compare the participants' submitted runs against human annotated, ground-truth memorability scores, and announce performance evaluation metrics for each participant's runs. The task has run for several years \cite{sweeney2022overview,savran2021overview,de2020overview} and has led to significant improvements in the performance of automatic memorability prediction for short form videos.

Memorability scores calculated for this work come from an an updated and adapted version of a video memorability prediction model presented in 2021 \cite{sweeney2021predicting}. The approach uses a Bayesian Ridge Regressor (BRR) to model the memorability prediction as regression task. The BRR model was trained on CLIP (Contrastive Language–Image Pre-training) features \cite{radford2021learning} extracted from the Memento10k training dataset. Given an input image, the model outputs a memorability scores ranging from 0 to 1, where a higher score indicates higher memorability. A single memorability score is generated for a video by averaging the memorability scores for a set of selected video frames.

\subsection{Image Saliency}

The concept of image saliency pertains to the extent to which an object or a specific region within an image or video frame differentiates itself from the surrounding elements in a visual scene \cite{wang2011image}. Essentially, it quantifies the capacity of a particular segment of a visual scene to capture the attention of a typical human observer. This concept is of paramount importance in the domains of computer vision and image processing, as it facilitates the strategic allocation of computational resources to regions deemed visually significant. Models of saliency endeavour to ascertain the regions of an image that are most likely to captivate human attention, and are thus instrumental in an array of applications, ranging from image compression and object recognition to visual search \cite{kummerer_deep_2015}.

The first DeepGaze saliency model was introduced in 2015. This work developed a novel approach to improving fixation prediction models by reusing existing neural networks trained on object recognition \cite{kummerer_deep_2015}. DeepGaze II used different feature spaces but the same readout architecture to predict human fixations in an image. This highlighted the importance of understanding the contributions of low and high-level features to saliency prediction \cite{kummerer_understanding_2017}. DeepGaze IIE \cite{linardos2021deepgaze}, the current version, showed that by combining multiple backbones in a principled manner, a good confidence calibration on objects in unseen image datasets can be achieved, resulting in a significant leap in benchmark performance with a 15\%  improvement over DeepGaze II reaching an AUC of 88.3\% on the MIT/Tuebingen Saliency Benchmark \cite{linardos2021deepgaze}.

Several studies have found a small correlations between visual salience and image memorability in specific contexts \cite{isola2011understanding,dubey2015makes,mancas2013memorability}. This relationship is strongest when images feature a single or limited number of objects presented in an uncluttered, close-up context. However, the introduction of multiple objects and fixation points considerably diminishes the association between the two \cite{dubey2015makes}, thereby signifying the distinctiveness of memorability and salience. Building upon this understanding, we leverage saliency to isolate specific sections of video frames. The underlying premise is that by magnifying the most salient part of a video frame---achieved by cropping the surrounding areas---the resultant cropped video, now more concentrated on the salient elements, could potentially enhance its memorability.

\subsection{The Memento10k Dataset and Memorability Scores}

%Newman {\it et al.} \cite{newman2020multimodal} worked with memorability of short form videos and introduced and published a dataset called Memento10k. The paper  discussed the reasons behind memorability
%decay in humans, which has implications in the
%real world. In the previous papers, memorability is explained
%based on visual and semantic annotations. In the Newman {\it et al.} paper   \cite{newman2020multimodal}, the authors extend
%the scope of memorability by introducing a decay into memorability over time.
%The concept of memory decay   can be traced back to the Ebbinghaus forgetting curve
%\cite{murre2015replication}, which was the result of experiments conducted in 1885. The Memento10K decay curve
%and Ebbinghaus forgetting curve suggest that memories are
%lost gradually over time but with a greater rate of forgetting
%in the first few days. 

In  work in this paper we use the  Memento10k dataset \cite{newman2020multimodal}, the most substantial video memorability dataset available for public use. The dataset as a whole comprises over 10,000 short form videos and nearly 1 million human annotations, providing a rich information source for analysing video media memorability.
The videos  cover diverse everyday events captured in a casual, homemade manner, enhancing the real-world applicability of the findings of those who use it. To ensure a robust evaluation process, we used a   subset of 1,500 videos from the dataset for  testing and evaluation , the same 1,500 videos as used in the evaluations in the MediaEval benchmark.

\begin{figure}[htbp]
    \centerline{\includegraphics[width=0.8\textwidth]{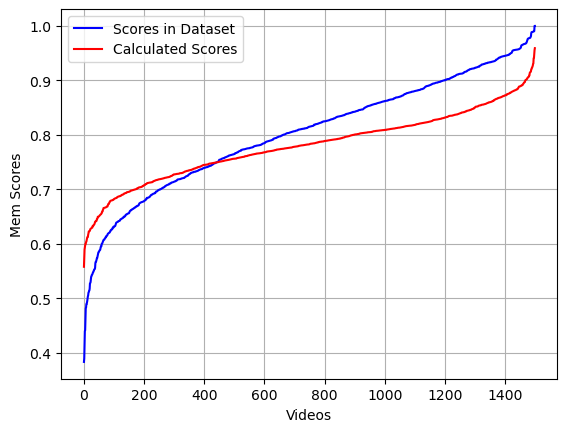}}
   \caption{Memorability scores calculated from \cite{sweeney2021predicting} vs. manually determined memorability scores in the Memento10k dataset provided for 1,500 test videos
   \label{fig:DataAnalysisGraph}}
\end{figure}

Each video in  Memento10k  has a  duration of approximately 3 seconds or 90 frames per video. We selected every 10th frame of each video for analysis to reduce computational complexity 
thus each video  has 9 representative frames. 
We used the model in \cite{sweeney2021predicting} to compute memorability scores for each of the 1,500 test videos.  The results of this are shown in 
Figure~\ref{fig:DataAnalysisGraph} which is a comparison between the ground truth of manually determined memorability scores provided with the dataset and  memorability scores generated by the model in \cite{sweeney2021predicting}. 
The manually determined memorability scores for the test videos  ranges from  0.38 to 0.99, where a lower score implies  lower memorability of a video and vice-versa.  Figure~\ref{fig:DataAnalysisGraph} shows the performance of the model to be quite accurate though it is more conservative  than  the ground truth.  For our work we  use the generated scores as a baseline for improving memorability by cropping. 

\begin{figure*}[!ht] % FIGURE environment 000
   \begin{center} % bringing picture
       \includegraphics[width=0.85\textwidth]{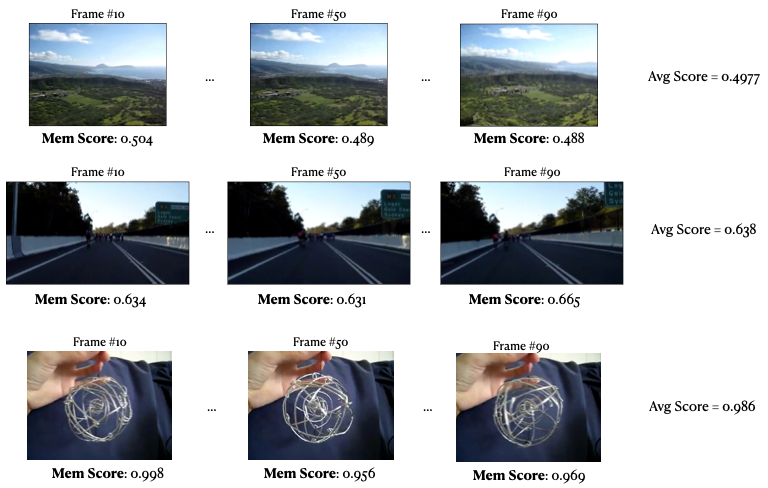}
   \end{center}
   \caption{Sample frames from 3 videos, memorability scores for those frames, and average memorability scores for the videos.
   \label{fig:sampleFrame} } % Label the image for later ref 
\end{figure*}

To illustrate, 
Figure~\ref{fig:sampleFrame} shows  sample frames from some videos, their corresponding memorability scores for individual frames, and the cumulative memorability score for each video.
We can see in Figure~\ref{fig:sampleFrame} that the first two videos, which seem more aesthetically pleasing, surprisingly exhibit lower memorability scores than the third video.

\section{Experimental Methodology}
    % \begin{itemize}
    %     \item Start with Lorin's Model \cite{sweeney2021predicting} and how we are computing memorability scores for a video.
    %     \item Explain how salience model (deepgaze 2e) is used to calculate saliency map and centre point.
    %     \item Explain salience based cropping.
    % \end{itemize}

\subsection{Predicting Video Memorability}

We used a vision transformer model fine-tuned on the task of predicting video memorability to predict memorability \cite{cummins2022analysing}. A quantitative investigation of the memorability of a popular crime-drama TV series, CSI was made through the use this model. The CLIP \cite{radford2021learning} pre-trained image encoder was used to train a Bayesian Ridge Regressor (BRR) on the Memento10k dataset. This work found that video shot memorability scores for lead cast members in the TV series heavily correlate with the characters' importance, considering on-screen time as importance.

\subsection{Saliency-Based Cropping}

When people observe an image without a specified task to complete, we do not focus the same level of intensity on each part of the image. Instead, attentive mechanisms direct our attention to the most important and relevant elements of an image \cite{cornia2018predicting}. The importance of  different parts of an image can be computed
yielding a  heatmap of importance,  a saliency map. This can assist in understanding the significance of different parts of an image.  Cropping a video around the centre of the saliency heatmap of  individual frames and moving that crop around the screen from  frame to frame, and/or changing the size of the crop as the saliency map changes, may increase the memorability of the overall video. Our rationale for employing frame cropping  is  eliminating noise from video frames and making the viewer focus on the most salient parts of the frame by removing the remainder.

We used a pre-trained DeepGaze IIE model \cite{linardos2021deepgaze} to compute the saliency map of video frames, the output of which is a saliency map with individual pixel values ranging from 0 to 1. The saliency map is denoted  $S$ and the within-frame location of the cropping center for a frame $(x,y)$ is calculated as the weighted center of all saliency values in the frame with coordinates given as:

\begin{equation}
    \begin{aligned}
        x = \frac{\sum_{i,j} i*S[i][j]}{\sum_{i,j} S[i][j]}
        ~~~~~~y = \frac{\sum_{i,j} j*S[i][j]}{\sum_{i,j} S[i][j]}
    \end{aligned}
\end{equation}

\noindent
We computed the area for frame cropping from the average and the standard deviation of the saliency map of every frame in a video, denoted as $a_{0}, a_{1},...a_{n}$ where $n$ is the number of frames. We then fit frame indices and used $\sqrt{a}$ for smoothing purposes to a linear function using a linear least least squares method. Videos with smaller fitting error were selected for linear zooming.

Linear zooming is a basic setup for zooming and involves changing the size of the frame cropping throughout the duration of a video. When we add a zoom to the cropping of a video we do so from the first to the last frame and in a liner fashion. We did not investigate variable crop zooming, or even zooming followed by reverse zoom within a video as this would be too disorientating for viewers because videos in Memento10k have short duration. 

We investigated three  approaches to combining saliency-based cropping and zooming.
In the first, we fixed the crop to the centre of the frame in order to allow us to measure the impact of zooming on memorability.  In the second approach we use tracking to follow the movement of the most salient part of frames while keeping the crop size fixed and the final approach allowed the crop size to change as the saliency moved and increased/decreased throughout the  video. 
We now describe each approach.

\vspace{-12pt}

\subsubsection{\textbf{1. Cropping at the video centrepoint:}}

Here we fixed the cropping of each video frame at the centrepoint and  analysed the changes in memorability scores resulting from this.
We systematically cropped  frames at percentage levels ranging from 10\% to 90\%, along a linear x- or y-axis, i.e. not by area. 
It is important to note that not all videos had  their salient part in the middle of the frame and thus this  should be considered a preliminary approach, laying the groundwork for more sophisticated video cropping.

\vspace{-12pt}

\subsubsection{\textbf{2. Tracking saliency with a fixed crop size:}}

\begin{figure*}[!ht]
  \begin{center}
      \includegraphics[width=0.9\textwidth]{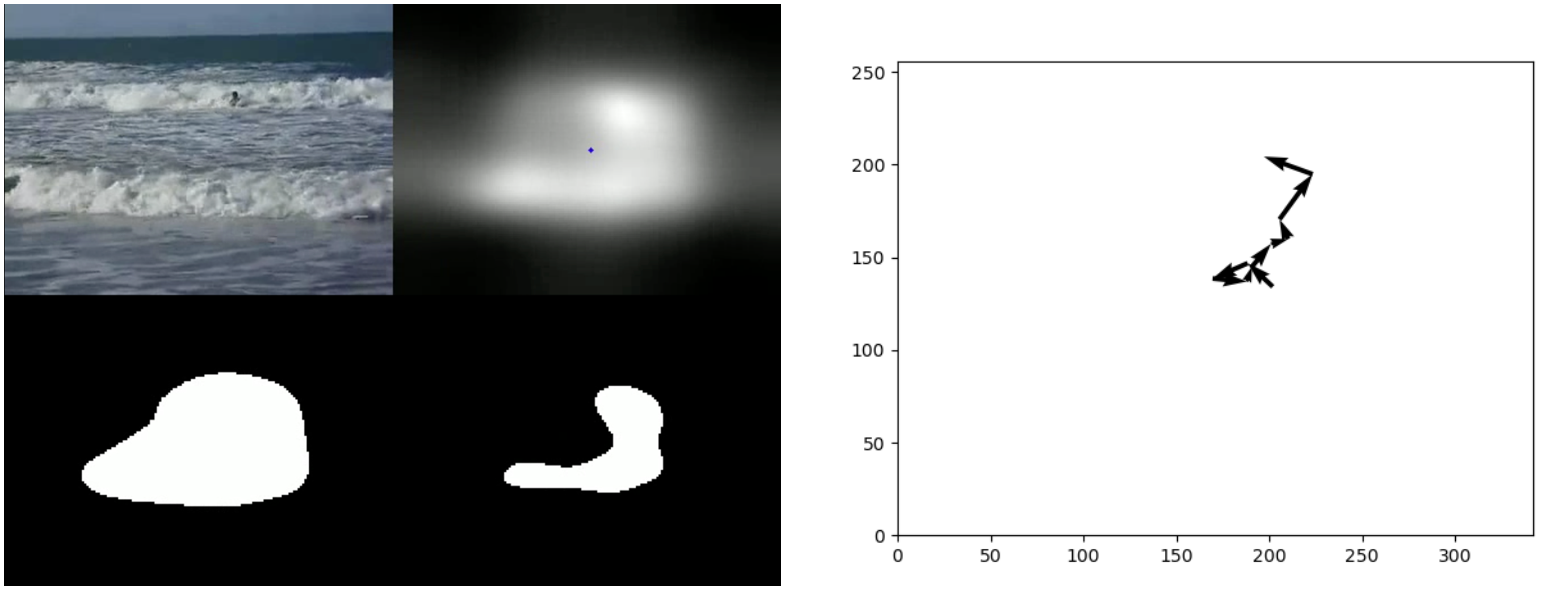}
  \end{center}
   \caption{Video frame (top left) and its generated saliency map with the centre point of the saliency spread marked as a point (top right). The image on the left also shows the saliency map at two different thresholds. The graph on the right shows the movement of the centrepoint of saliency for the duration of the video.
   \label{SaliencyCentre}}
\end{figure*}

\begin{figure*}[!h]
  \begin{center}
      \includegraphics[width=\textwidth]{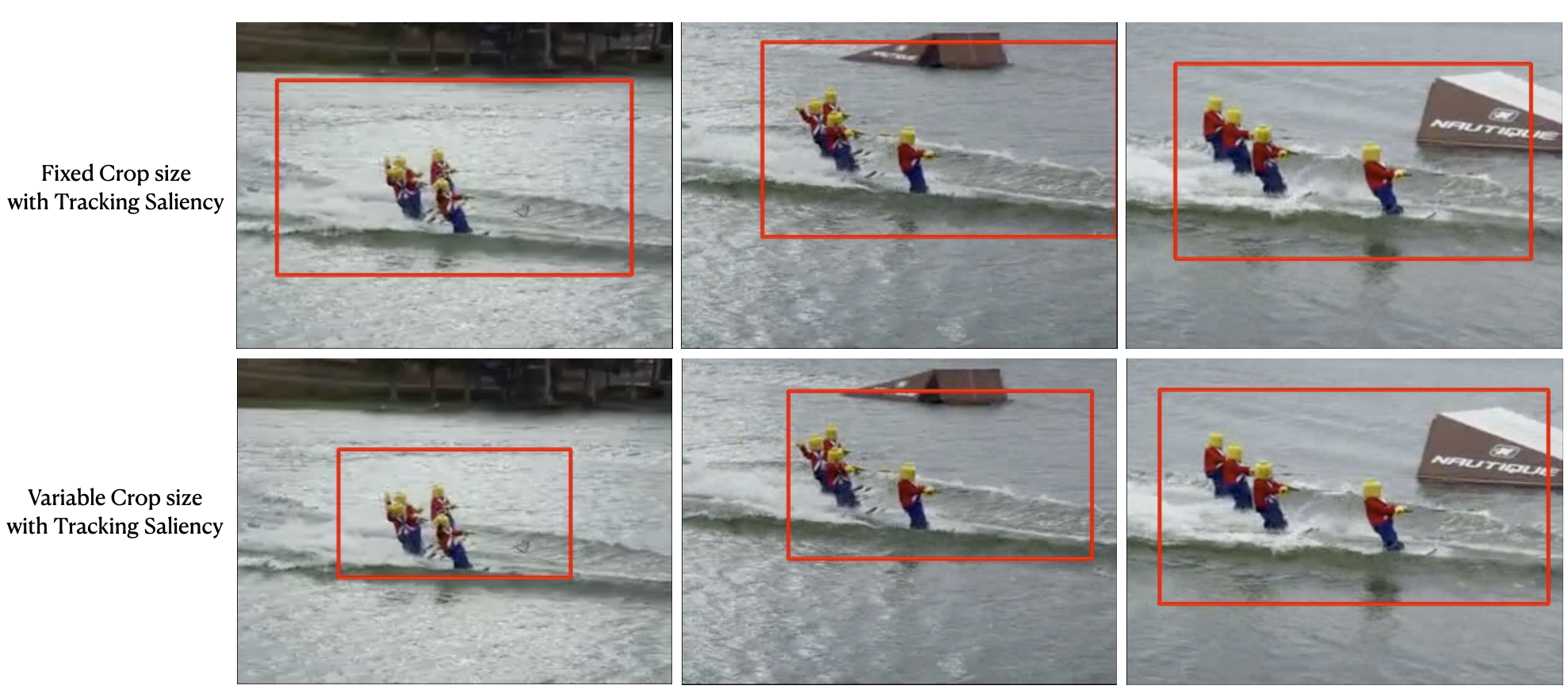}
  \end{center}
   \caption{Sample frames for fixed-sized (second approach) and for variable-sized cropping (third approach) with saliency tracking.
   \label{Exampleframe}}
\end{figure*}

In the second approach we determined the centrepoints of the saliency from the output of the DeepGaze IIE saliency model \cite{linardos2021deepgaze} for each frame.  We then tracked their movement around the frame throughout each video while maintaining a fixed size for the bounding box surrounding this region. The saliency map was generated at different thresholds in order to binarise it as shown below and the results at different saliency thresholds are shown in the bottom-half of Figure~\ref{SaliencyCentre}. 
The thresholds used are shown below and an illustration of this cropping is shown in the top sequence of frames in Figure~\ref{Exampleframe}.

\begin{gather}
  \emph{\text{saliency\_map}} > \emph{\text{saliency\_mean} + \text{saliency\_std}} \\
  \emph{\text{saliency\_map}} > \emph{\text{saliency\_mean}} + \emph{2}\times\emph{\text{saliency\_std}}
\end{gather}

\noindent

\vspace{-12pt}

An illustrative example of the resulting frames can be observed in the top row of images corresponding to three frames from a video in Figure~\ref{Exampleframe}. It can be seen in this example  that the size of the bounding box shown in red remains constant throughout the video.
The main limitation of fixing the bounding box is that it does not consider any changes in the spread or contraction of the most salient part of the frames as the video progresses and might allow either a non-salient part in the cropped frame or might crop out a salient part.

\setlength{\belowcaptionskip}{-10pt}

\noindent
In practice, the thresholding condition forces the viewer to focus on areas of the frame with higher saliency values corresponding to regions of interest compared to the background noise or the less visually significant regions.
It is important to note that the spread of the saliency  within the frame might vary over time, either increasing and spreading or reducing and contracting, thus introducing dynamic changes in the salient region size. However, the fixed cropping size used for the bounding box in this approach did not account for  temporal variations in saliency spread.

\subsubsection{\textbf{3. Tracking saliency with a variable crop size:}}

In the third approach,  tracking  involved monitoring the salient part with the crop size dynamically adjusted by linearly increasing, or decreasing, or neither, based on the size of the identified salient region.

An example of the resulting frame for this approach is shown in the second row of images in  Figure~\ref{Exampleframe}. Here it can be seen that the size of the bounding box  increases as  the video progresses.
To accommodate changes in the spread or decrease in saliency throughout the video, only videos with a consistent increase, or decrease, in thresholded saliency size were cropped linearly thus taking changes in saliency spread into consideration.
Finally, in all cases where we used  cropping we incorporated padding around the crop in order to  provide some context for the most salient parts of the video frames.

\section{Experimental Results}

\subsection{Cropping at Video Centrepoints}
        
\begin{figure}[ht]
            \centerline{
            \includegraphics[width=0.55\textwidth]{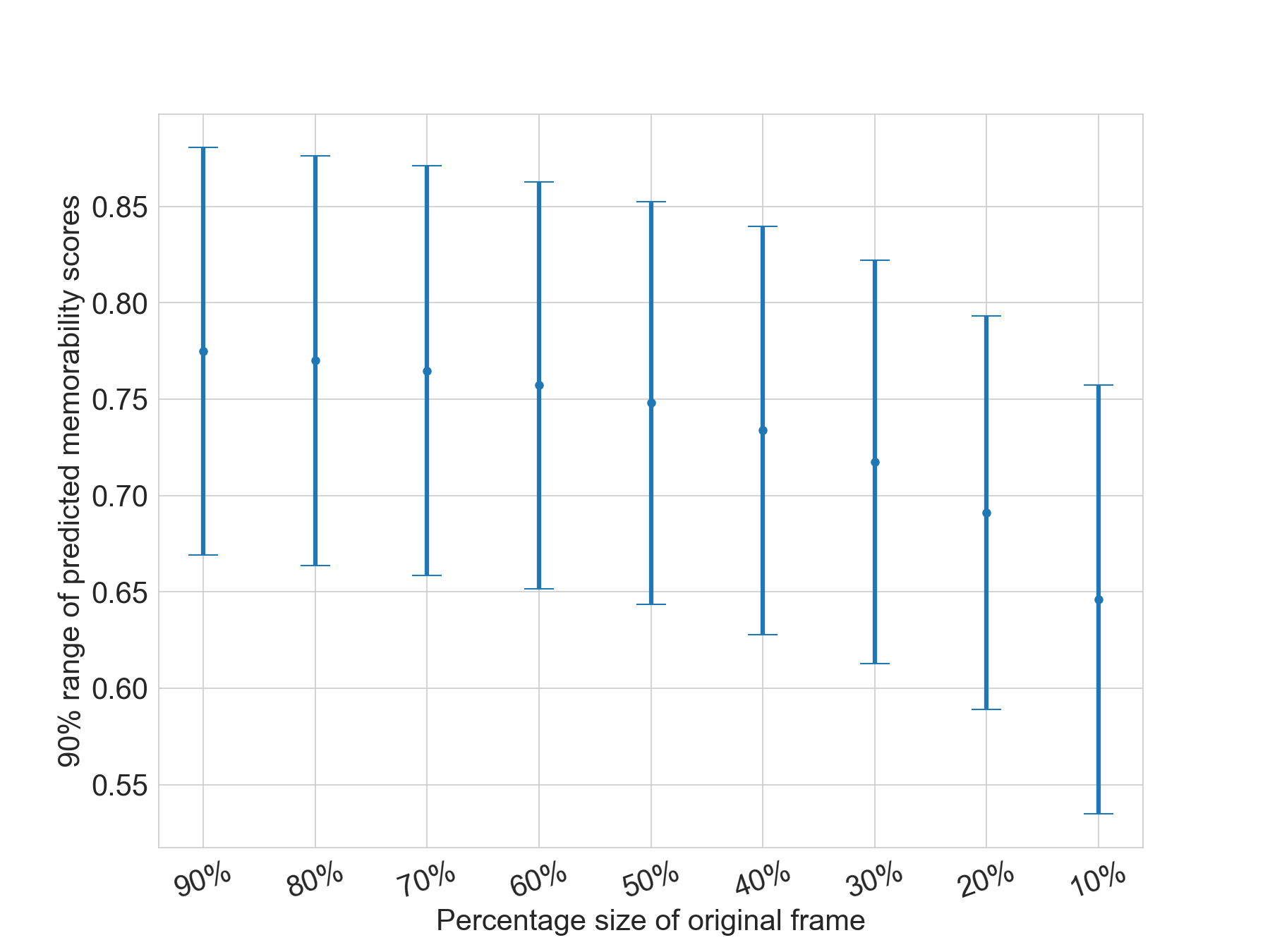}
             \raisebox{0.15\height}{\includegraphics[width=0.42\textwidth]{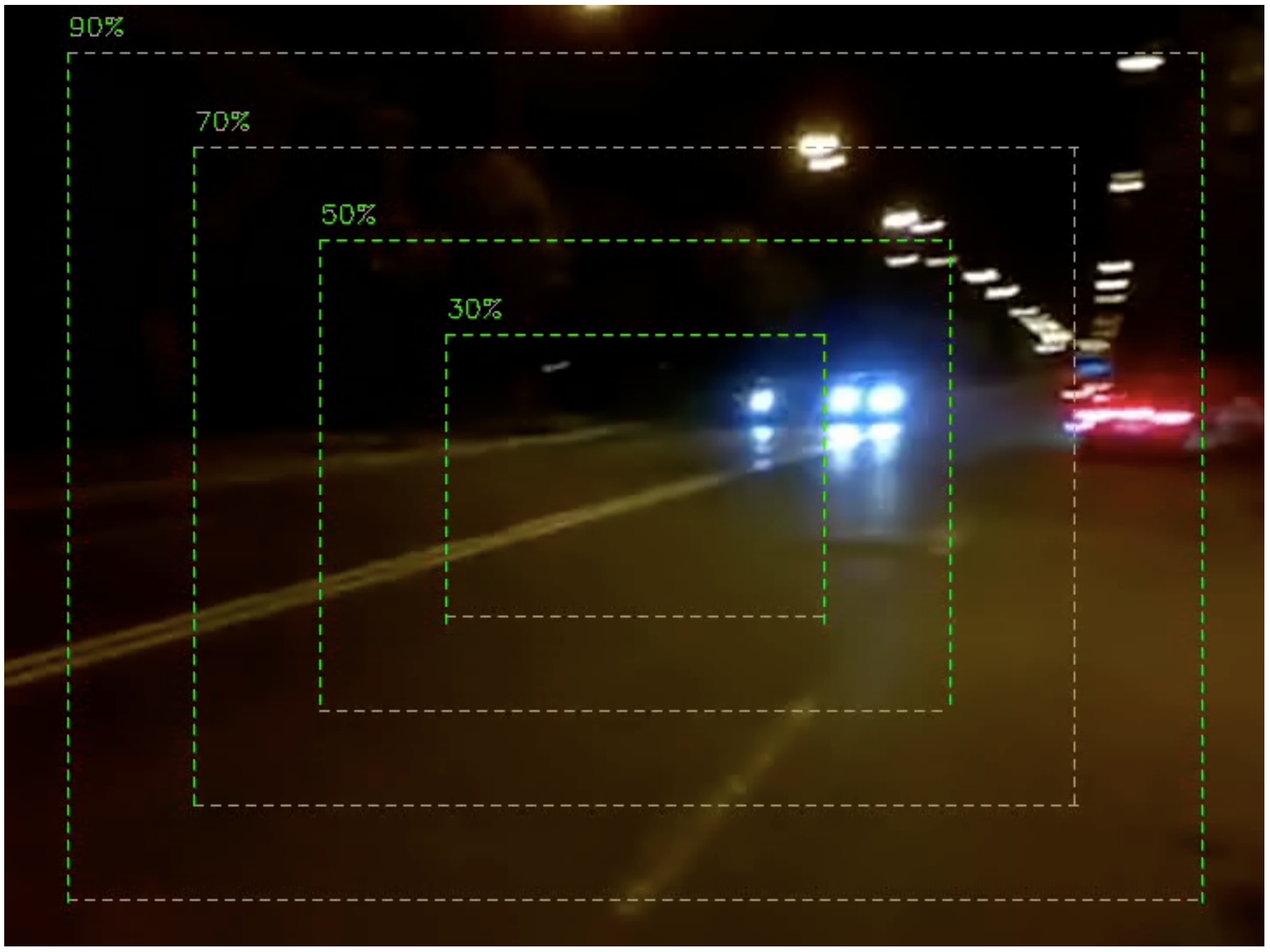}}
             }
           \caption{Changes in predicted video memorability as a result of varying crop sizes where a crop size of 90\% means discarding 10\% of the frame}
           \label{CentralCrop}
        \end{figure}

The range plot on the left in Figure~\ref{CentralCrop} shows the range for 90\% of the new  memorability prediction scores from our initial centrepoint cropping across all 1,500 test videos.  The image on the right   illustrates that  cropping. As we  reduce the  sizes  of the frames by cropping from 90\% down to 10\% of their original sizes, the 90\% ranges of memorability scores show a noticeable decrease. This is probably caused by  cropping ignoring the within-frame locations of the most salient parts of the frame and most likely cropping through, rather than around, objects of interest. This result justifies a further exploration into a more saliency-informed approach to cropping.

\setlength{\belowcaptionskip}{-10pt}

\subsection{Cropping with Saliency Tracking}
    
\begin{figure*}[!ht]
\centering
\includegraphics[width=0.9\textwidth]{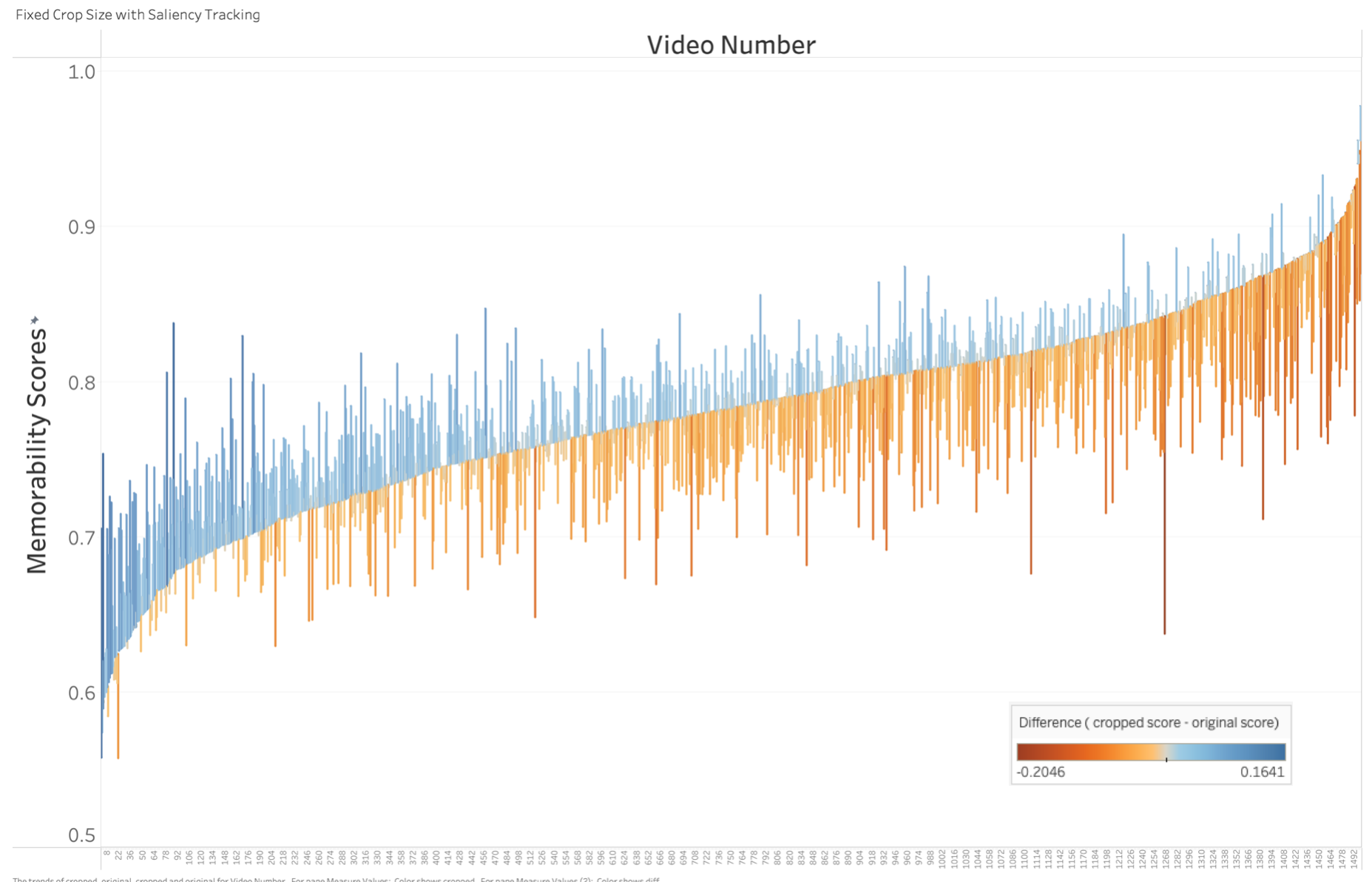}\\
\includegraphics[width=0.9\textwidth]{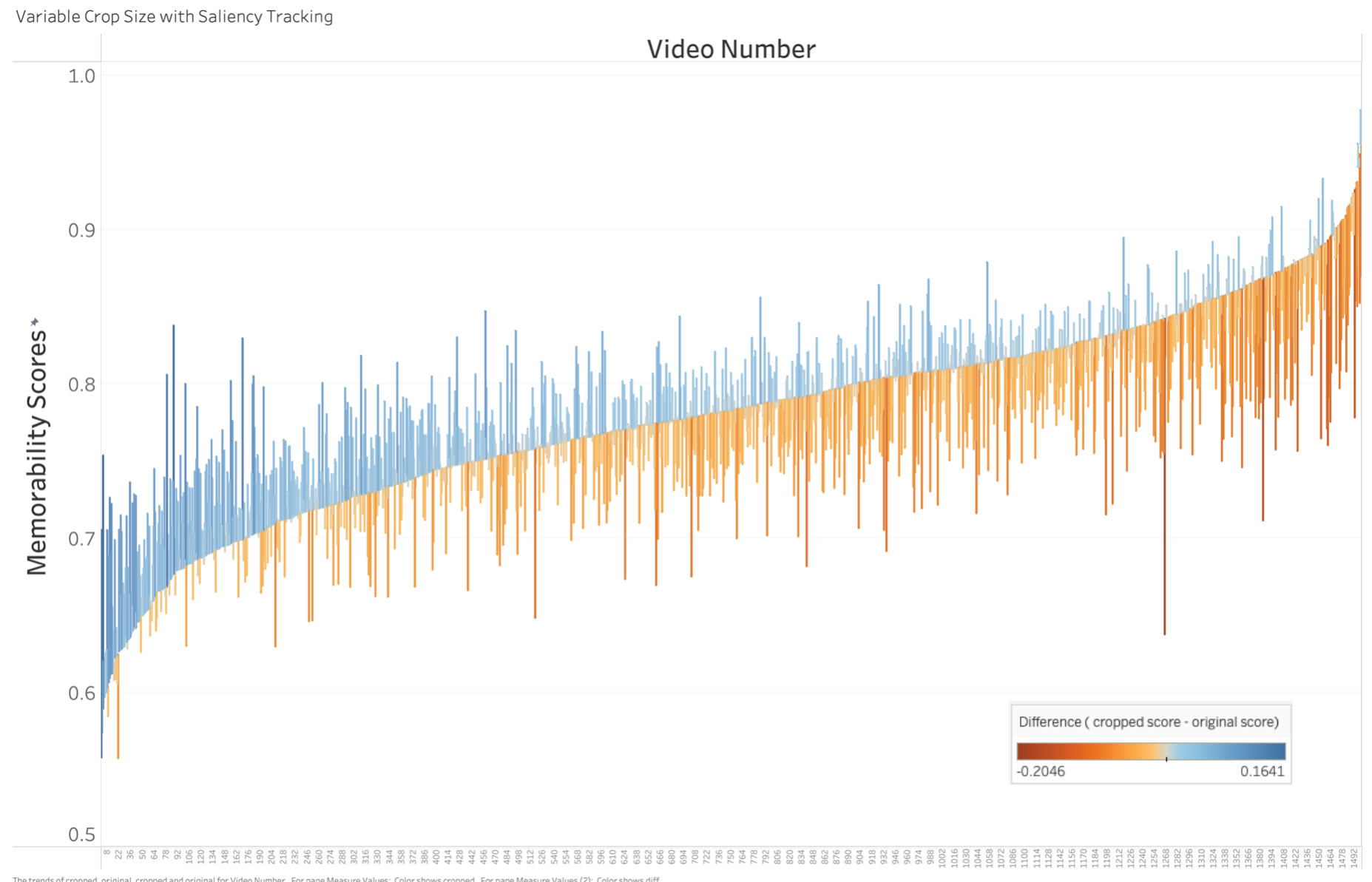}
\caption{Results for changes in predicted memorability from fixed crop size (top) and variable crop size (bottom), both with saliency tracking. Blue lines  show where scores after cropping are improvements on the original memorability score and orange lines are the opposite. \label{fixedVarcrop}
}
\end{figure*}

The graphs in Figure~\ref{fixedVarcrop} show  results when comparing predicted memorability scores after cropping with scores for the original 1,500 videos for both fixed (top graph) and variable (bottom graph) crop sizes. Videos in the graphs are sorted left to right by increasing values of their initial memorability scores. Blue lines show where  cropping improved the  score,  orange lines show where it was reduced.

\setlength{\belowcaptionskip}{-10pt}

For each graph it can be seen that for low initial memorability scores blue lines tend to dominate, and this reverses  as  scores increase with orange lines dominating. 
For the fixed crop size with tracking, 707 videos have improved memorability prediction scores while 794 have decreased scores. When we introduce variable crop sizes with tracking we find that 718 videos have improved scores with 783 with decreased and this explains why the graphs are so similar, the variable crop size did not have much impact on the results.  The distribution of these changes in scores for variable crop sizes with tracking is summarised in Table~\ref{tab:distribs} where we see 83.1\% of videos with an initial score above 0.7 have improvements, falling to less than half when the threshold is 0.9.

\vspace{-12pt}

\begin{table}[]
    \centering
        \caption{Distribution of videos with improved scores from variable sized cropping by initial memorability scores.}
    \begin{tabular}{lllllll}
    \toprule
         Threshold memorability score& 0.70 & 0.75 & 0.80 & 0.85 & 0.90 & 0.95 \\
         Videos with improved scores~~~~ & 83.1\%~~~ & 73.0\%~~~ & 58.8\%~~~ & 51.7\%~~~ & 45.7\%~~~ & 47.8\% \\
         \bottomrule
    \end{tabular}
    \label{tab:distribs}
\end{table}

\vspace{-12pt}

One explanation for this is that for videos which are more memorable, cropping removes memorable information or at best it removes some of the context for memorable information in the video frame thus reducing the video's overall memorability.
For videos which already have lower memorability scores, cropping generates improved memorability because it removes visual noise from the frames, allowing the viewer to concentrate on the most salient, and probably more memorable, aspects of the videos.

%\begin{figure}[htbp]
%    \centerline{\includegraphics[width=0.5\textwidth]{pic/PercentImprovedThreshold.png}}
%   \caption{Percentage of videos with improved   memorability scores by cropping using saliency for videos with a threshold memorability score
%   \label{ImprovementThreshold}}
%\end{figure}

\begin{figure}[htbp]
    \centerline{\includegraphics[width=0.85\textwidth]{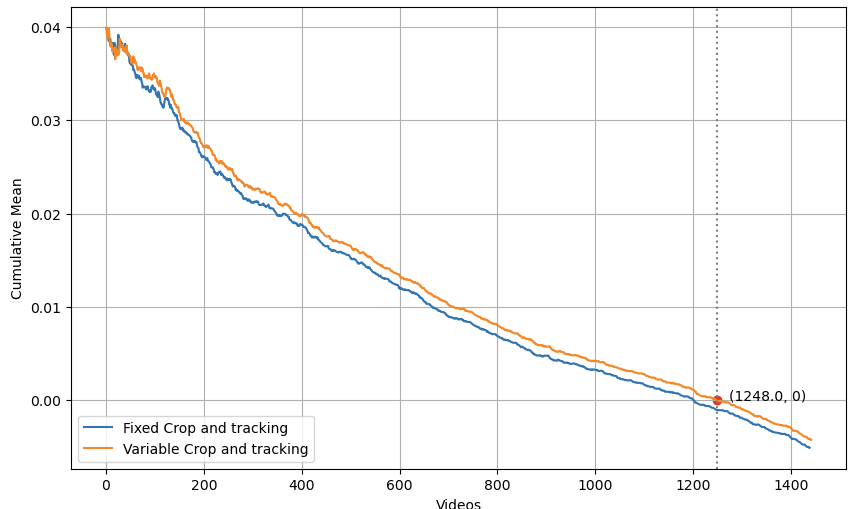}}
   \caption{Line graph showing score comparison between fixed and variable cropping -- the x-axis shows  1,500 videos and the y-axis is the cumulative mean of changes in memorability score.
   \label{Metric}}
\end{figure}

%
%  \begin{figure}[htbp]
%     \centerline{\includegraphics[width=0.35\textwidth]{pic/Boxplot_outliers.png}}
%    \caption{Outliers in the cumulative mean data}
%    \label{Box_outliers}
% \end{figure}
%
This interpretation of the results can be shown by the graph in Figure~\ref{Metric} which is the distribution of the ``cumulative mean" for memorability score changes as a result of both fixed and variable cropping of  videos. In calculating this, the cumulative mean at step $i$ = $\frac{\sum_{k=1}^{i} x_k}{i}$ where $x_k$ represents the data point at step $k$ in the sequence and $i$ is the current step for which the cumulative mean is being calculated.  We removed  outliers from this using the inter-quartile method to make an unbiased plot.

There are two notable conclusions from Figure~\ref{Metric} which are that the variable cropping method (orange line) performs slightly better than the fixed cropping method and that the cumulative mean decreases as the original memorability score increases for a video. Hence we can say that cropping improves  memorability scores for most  videos with an original memorability score lower than a  threshold and beyond that, cropping  starts to remove  salient part of the frames.

%   \begin{figure*}[htbp]
%   \begin{center}
%       \includegraphics[width=0.8\textwidth]{pic/FixCropTrack.png}
%   \end{center}
%    \caption{Results for \textbf{Fixed crop size} with saliency tracking. (The blue part in the graph shows that the mem score after cropping is better than the original mem score and vice-versa.)}
%    \label{fixedcrop}
% \end{figure*}

%   \begin{figure*}[htbp]
%   \begin{center}
%       \includegraphics[width=0.8\textwidth]{pic/VarCropTrack.png}
%   \end{center}
%    \caption{Results for \textbf{Variable crop size} with saliency tracking. (The blue part in the graph shows that the mem score after cropping is better than the original mem score and vice-versa.)}
%    \label{VariableCrop}
% \end{figure*}

\section{Conclusions}

The investigation outlined in this paper explores the potential of improving video memorability prediction by applying frame cropping based on visual saliency. Using a state-of-the-art saliency model, DeepGaze IIE \cite{linardos2021deepgaze}, we detected and tracked saliency throughout a video, cropped the frames around the most salient parts, and re-generated the video. Our results on the Memento10k test set \cite{newman2020multimodal} demonstrate a notable improvement in memorability prediction for the majority of videos with initial memorability scores up to a certain threshold.

While this approach is undoubtedly effective for videos with lower memorability scores---eliminating noise or less salient parts of a video---it does have limitations. Videos already exhibiting high memorability scores are likely to have inherently lower levels of noise and a more prominent narrative. In such cases, exploring more sophisticated visual manipulations, such as temporal segmentation (selectively retaining contextually important segments while removing less crucial/incoherent ones), employing advanced object recognition models to identify and alter key elements within the frame with video in-painting \cite{Kim_2019_CVPR}, or applying techniques like video super-resolution \cite{liu2022learning} to improve the quality of the videos, is necessary. Furthermore, employing dynamic scene composition adjustments to better highlight key actions or objects in the video may also prove fruitful. Ultimately, a multifaceted approach that explores various advanced visual manipulation techniques will be necessary to optimise the memorability of videos which already manifest high memorability scores.

%\subsection*{Acknowledgements}
%We thank whoever you want to thank.

%%%%%%%%%%%%%%%%%%%%%%%%%%%%%%%%%%%%%%%%%%%%%%%%%%%
%Bibliographies: First load the bibTex file...from citation 
\bibliographystyle{plain} % style
\bibliography{bibliography.bib} % say bib-file name
\end{document}